\documentclass{SCIS2026}
\usepackage{algorithm}
\usepackage{float}
\usepackage{makecell} 
\usepackage{booktabs} 
\usepackage{placeins} 
\usepackage{subcaption}
\begin{document}
\ArticleType{RESEARCH PAPER}
\Year{2025}
\Month{January}
\Vol{68}
\No{1}
\DOI{}
\ArtNo{}
\ReceiveDate{}
\ReviseDate{}
\AcceptDate{}
\OnlineDate{}
\AuthorMark{}
\AuthorCitation{}

\title{DisCo3D: Distilling multi-view consistency for 3D scene editing}{Chi Y F, Ma H M, Wang K F, et al. DisCo3D: distilling multi-view consistency for 3D scene editing}

\author[1]{Yufeng CHI}{}
\author[2]{Huimin MA}{}
\author[1]{Kafeng WANG}{}
\author[1]{Jianmin LI}{{lijianmin@mail.tsinghua.edu.cn}}

\address[1]{Institute for Artificial Intelligence, Beijing National Research Center for Information Science and Technology,\\ Department of Computer Science and Technology, Tsinghua University, Beijing 100084, China}
\address[2]{College of Computer, National University of Defense Technology, Changsha 410073, China}

\abstract{While diffusion models have demonstrated remarkable progress in 2D image generation and editing, extending these capabilities to 3D editing remains challenging, particularly in maintaining multi-view consistency. Classical approaches typically update 3D representations through iterative refinement based on a single editing view. However, these methods often suffer from slow convergence and blurry artifacts caused by cross-view inconsistencies. Recent methods improve efficiency by propagating 2D editing attention features, yet still exhibit fine-grained inconsistencies and failure modes in complex scenes due to insufficient constraints. To address this, we propose \textbf{DisCo3D}, a novel framework that distills 3D consistency priors into a 2D editor. Our method first fine-tunes a 3D generator using multi-view inputs for scene adaptation, then trains a 2D editor through consistency distillation. The edited multi-view outputs are finally optimized into 3D representations via Gaussian Splatting. Experimental results show DisCo3D achieves stable multi-view consistency and outperforms state-of-the-art methods in editing quality.}


\maketitle

\section{Introduction}

\begin{figure}[t]
  \centering
  \includegraphics[width=0.95\linewidth]{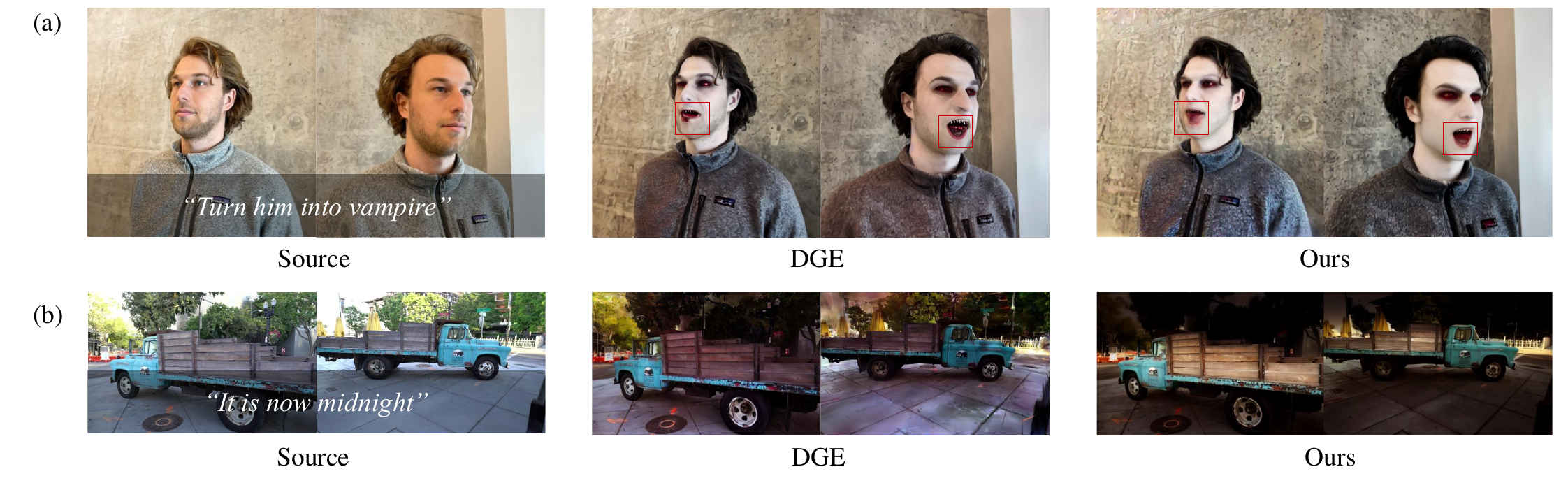}
  \caption{
  \textbf{Comparison of multi-view consistent scene editing.} We compare our method against a representative baseline, DGE~\cite{chen2024dge}.
  (a) \textit{Fine-grained inconsistencies}: The baseline suffers from residual fine-grained inconsistencies, failing to align high-frequency details across views (e.g., vampire's mouth in red box).
  (b) \textit{Failure modes in complex scenes}: Under large viewpoint variations (``It is now midnight'' edit), the baseline exhibits failure modes associated with progressive error accumulation. In contrast, our method avoids artifact accumulation through consistency distillation.
  }
  \label{fig:intro}
\end{figure}

Recent advances in diffusion models have revolutionized 2D image generation and editing~\cite{ho2020denoising,song2020denoising,brooks2023instructpix2pix,zhang2023adding,zhang2023sine}, enabling high-quality results with precise semantic control. However, extending these capabilities to 3D scene editing remains challenging due to inherent multi-view consistency requirements. Classical approaches~\cite{haque2023instruct,chen2024gaussianeditor,xu2023instructp2p,wang2024proteusnerf,yu2023edit} employ Iterative Dataset Update (IDU) frameworks that cyclically refine 3D representations based on edits from a single view, but they suffer from slow convergence and blurry artifacts due to conflicting signals across viewpoints.
Recent methods~\cite{chen2024dge,dong2023vica,gomel2024diffusion,wang2024view,wu2024gaussctrl,kwon2024unified,lee2025editsplat} attempt to enforce multi-view consistent editing through two predominant paradigms: either through cross-view attention feature propagation or via depth-map conditioned control, both operating under implicit editing space constraints. However, residual fine-grained inconsistencies persist across edited views, manifesting as misaligned high-frequency details (e.g., vampire's mouth textures in Figure~\ref{fig:intro} (a)). These artifacts arise from insufficient regularization of high-frequency details during feature alignment. Moreover, the challenge intensifies when handling complex scenes with significant viewpoint variations, where current attention-based methods exhibit distinct failure modes (Figure~\ref{fig:intro} (b)): key-view-based methods~\cite{chen2024dge,wu2024gaussctrl} suffer progressive error accumulation when propagating attention features across geometrically discontinuous regions, and single-reference guided techniques~\cite{gomel2024diffusion,kwon2024unified} degrade under severe occlusion due to incomplete contextual constraints. Furthermore, all-view attention integration~\cite{wang2024view} compromises fidelity when resolving inter-view conflicts. These limitations prevent existing frameworks from achieving stable multi-view consistency, highlighting the need for a unified architecture that enforces fine-grained consistency while robustly reconciling macroscopic view relationships in challenging scene configurations.

To address these, we propose \textbf{DisCo3D}, a novel framework that distills consistency priors from 3D scene generators into 2D editing models, enabling precise multi-view consistent editing with direct 3D representation updates.  Our key insight lies in leveraging the strong consistency awareness of 3D diffusion models to constrain the 2D editor's output space while maintaining semantic alignment with editing instructions. Specifically, DisCo3D first enhances a 3D novel view synthesis (NVS) diffusion model (e.g., \textit{ViewCrafter}~\cite{yu2024viewcrafter}) through Low-rank adaptation (LoRA)~\cite{hu2022lora} fine-tuning along predefined camera trajectories. Subsequently, we distill the consistency priors into an instruction-based 2D editor (e.g., \textit{InstructPix2Pix}~\cite{brooks2023instructpix2pix}) through a novel training objective combining distillation and regularization losses. The distilled editor generates multi-view consistent edits simultaneously, which are subsequently integrated into the explicit 3D Gaussian Splatting (GS)~\cite{kerbl20233d} representation.

Our contributions are as follows:
we propose a consistency distillation framework that transfers strong 3D consistency priors to a 2D editor, enabling efficient and fine-grained multi-view consistent editing. Additionally, we introduce a regularization strategy that balances semantic fidelity and geometric consistency, preserving the editor’s native capabilities and preventing undesired distortions. Our multi-view consistent editing results are directly integrated into the 3D GS representation, eliminating the need for iterative refinement. We conduct extensive comparisons across diverse datasets (\textit{IN2N}~\cite{haque2023instruct}, \textit{Tanks-and-Temples}~\cite{knapitsch2017tanks}), DisCo3D consistently outperforms prior methods across both quantitative and qualitative evaluations, demonstrating strong stability and effectiveness, particularly in complex 360-degree scenes where prior methods suffer from severe artifacts or semantic misalignment.
\label{sec:intro}

\section{Related work}

\subsection{Novel view synthesis (NVS) diffusion models}

Our approach relies on the multi-view consistency priors provided by NVS diffusion models, which extend the powerful image generation capabilities of diffusion models to synthesize novel views from a single or sparse set of input views.

Several distillation-based methods~\cite{wang2023prolificdreamer,poole2022dreamfusion,wang2023sjc} train 3D representations under the supervision of diffusion models, but their reliance on limited 3D data leads to suboptimal geometric fidelity. Zero-1-to-3~\cite{liu2023zero} addresses this by fine-tuning a diffusion model on large synthetic datasets~\cite{deitke2023objaverse,deitke2023objaversexl} with camera pose conditioning, significantly improving view synthesis quality. However, these models still struggle with complex backgrounds and cluttered scenes. Recently, ZeroNVS~\cite{sargent2024zeronvs} refines the camera pose conditioning and incorporates real-world data~\cite{reizenstein2021common,yu2023mvimgnet,zhou2018stereo}, enabling NVS with complex backgrounds. Cat3D~\cite{gao2025cat3d} adopts video diffusion architectures~\cite{ho2022video}, conditioning on camera ray representations to generate consistent multi-view outputs simultaneously. In contrast, NVS-Solver~\cite{you2024nvs} introduces a training-free paradigm by integrating depth-based warping constraints with pretrained video diffusion models~\cite{blattmann2023stable}. ViewCrafter builds upon DynamiCrafter~\cite{xing2024dynamicrafter}, an image-to-video diffusion model, and enhances pose control using the dense stereo model DUSt3R~\cite{wang2024dust3r}, achieving high-fidelity synthesis across general scenes.

Considering both model capabilities and open-source availability, we select ViewCrafter’s pre-trained model as the teacher model in our approach.

\subsection{Diffusion-based 3D scene editing}

Fine-grained control over 3D editing remains constrained by limited 3D data availability, often relying on 2D diffusion models. Several studies~\cite{cheng2023progressive3d,zhuang2023dreameditor,kamata2023instruct3dto3d,park2023ed} follow the Score Distillation Sampling (SDS) approach from DreamFusion~\cite{poole2022dreamfusion}, optimizing 3D representations~\cite{mildenhall2021nerf,kerbl20233d} using diffusion priors for editing, but typically produce blurred texture details. Instruct-NeRF2NeRF~\cite{haque2023instruct} introduces Iterative Dataset Update (IDU), which can be regarded as a variant of SDS, enabling 3D edits by modifying rendered views and iteratively updating the scene representation. Subsequent works including InstructP2P~\cite{xu2023instructp2p}, ProteusNeRF~\cite{wang2024proteusnerf}, EditNeRF~\cite{yu2023edit}, and GaussianEditor~\cite{chen2024gaussianeditor} all adopt this strategy, yet they exhibit slow convergence and blurred artifacts due to cross-view signal conflicts. 
GaussEdit~\cite{shu2025gaussedit} extends SDS-based editing by introducing ROI-level control on 3D GS and alternates between Stable Diffusion and a multi-view model with category-guided regularization to optimize Gaussian parameters. Despite a 3D-aware texture refinement stage, it still suffers from smooth textures due to SDS limitations and view inconsistency.
Recent works parallel to ours focus on multi-view consistent editing, DGE~\cite{chen2024dge} treats multi-view sequences as videos for joint editing with explicit constraints, propagating attention feature from key views to reduce inconsistency. However, such methods tend to accumulate alignment errors during feature propagation across geometrically discontinuous regions. Other approaches~\cite{gomel2024diffusion,kwon2024unified} guide multi-view editing by transferring features from reference views, but incomplete constraint propagation persists due to missing contextual information. Similar existing methods, including ViCA-Nerf~\cite{dong2023vica}, GaussCtrl~\cite{wu2024gaussctrl}, VCEdit~\cite{wang2024view} and InterGSEdit~\cite{wen2025intergsedit} achieve editing consistency through multi-view attention feature interactions or depth-based control mechanisms, yet insufficient constraints lead to residual fine-grained inconsistencies that ultimately propagate into 3D representations. Apart from these attention-based strategies, other perspectives have also been explored: ConsistDreamer~\cite{chen2024consistdreamer} introduces a 3D-consistent 2D diffusion framework via structured noise construction, while Chat-Edit-3D~\cite{fang2024chat} focuses on interactive editing workflows facilitated by Large Language Models. However, ConsistDreamer relies on heuristic structured noise initialization which may struggle to guarantee strict geometric consistency across large viewpoint changes, and Chat-Edit-3D is limited by the artifacts inherent in texture atlas parameterization. In contrast, our DisCo3D  addresses these limitations by explicitly distilling strong 3D consistency priors from a fine-tuned NVS teacher, while preserving semantic alignment through regularization, enabling stable multi-view consistency for high-quality 3D editing.

\section{Preliminaries}

\noindent\textbf{3D NVS diffusion model.}  
Based on video diffusion models~\cite{xing2024dynamicrafter}, frameworks like ViewCrafter~\cite{yu2024viewcrafter} employ 3D U-Nets to denoise Gaussian noise sequences, generating novel view sequences conditioned on reference images. The reference image \( I^{\text{ref}} \) is encoded as a CLIP~\cite{radford2021learning} embedding for cross-attention guidance. Additionally, a dense stereo module (e.g., DUSt3R~\cite{wang2024dust3r}) reconstructs the point cloud from \( I^{\text{ref}} \), implicitly incorporating camera trajectory information to render novel view maps, which are channel-wise concatenated with the sampled noise.

During training, DUSt3R simultaneously predicts camera parameters from multi-view ground truth sequences and generates render maps from the reference image \( I^{\text{ref}} \). The ground truth set and render maps are encoded into latent representations \( \mathbf{z_0} \) and \( \mathbf{\hat{z}} \), respectively. The 3D denoising model is optimized using the diffusion loss~\cite{ho2020denoising}:  
\begin{equation}\label{eq:vc_loss}
\min_{\theta} \mathbb{E}_{t \sim U(0, 1), \epsilon \sim \mathcal{N}(0, \mathbf{I})} 
\left[ \| \epsilon_\theta(\mathbf{z}_t, t, \mathbf{\hat{z}}, I^{\text{ref}}) - \epsilon \|_2^2 \right],
\end{equation}  
where $\mathbf{z}_t = \alpha_t \mathbf{z_0} + \sigma_t \epsilon$, in which $\alpha_t$ and $\sigma_t$ denote the noise schedule~\cite{karras2022elucidating} coefficients.

\noindent\textbf{Low-rank adaptation.}  
Low-rank adaptation (LoRA)~\cite{hu2022lora} provides an efficient approach for fine-tuning large language models by modifying only a small subset of parameters, making it well-suited for task-specific adaptation. Recently, this technique extends naturally to video models, enabling applications in video editing tasks such as Motiondirector~\cite{zhao2024motiondirector} and I2VEdit~\cite{ouyang2024i2vedit}. Given a pre-trained weight matrix \( \mathbf{W_0} \in \mathbb{R}^{d \times k} \), LoRA constrains its update using a low-rank decomposition:  
\begin{equation}
    \mathbf{W_0} + \Delta\mathbf{ W} = \mathbf{W_0} + \mathbf{B}\mathbf{A},
\end{equation}  
where \( \mathbf{B} \in \mathbb{R}^{d \times r} \) and \( \mathbf{A} \in \mathbb{R}^{r \times k} \), with rank \( r \) significantly smaller than \( d \) and \( k \).  

\section{Method}
\begin{figure}[t]
    \centering
    \includegraphics[width=\linewidth]{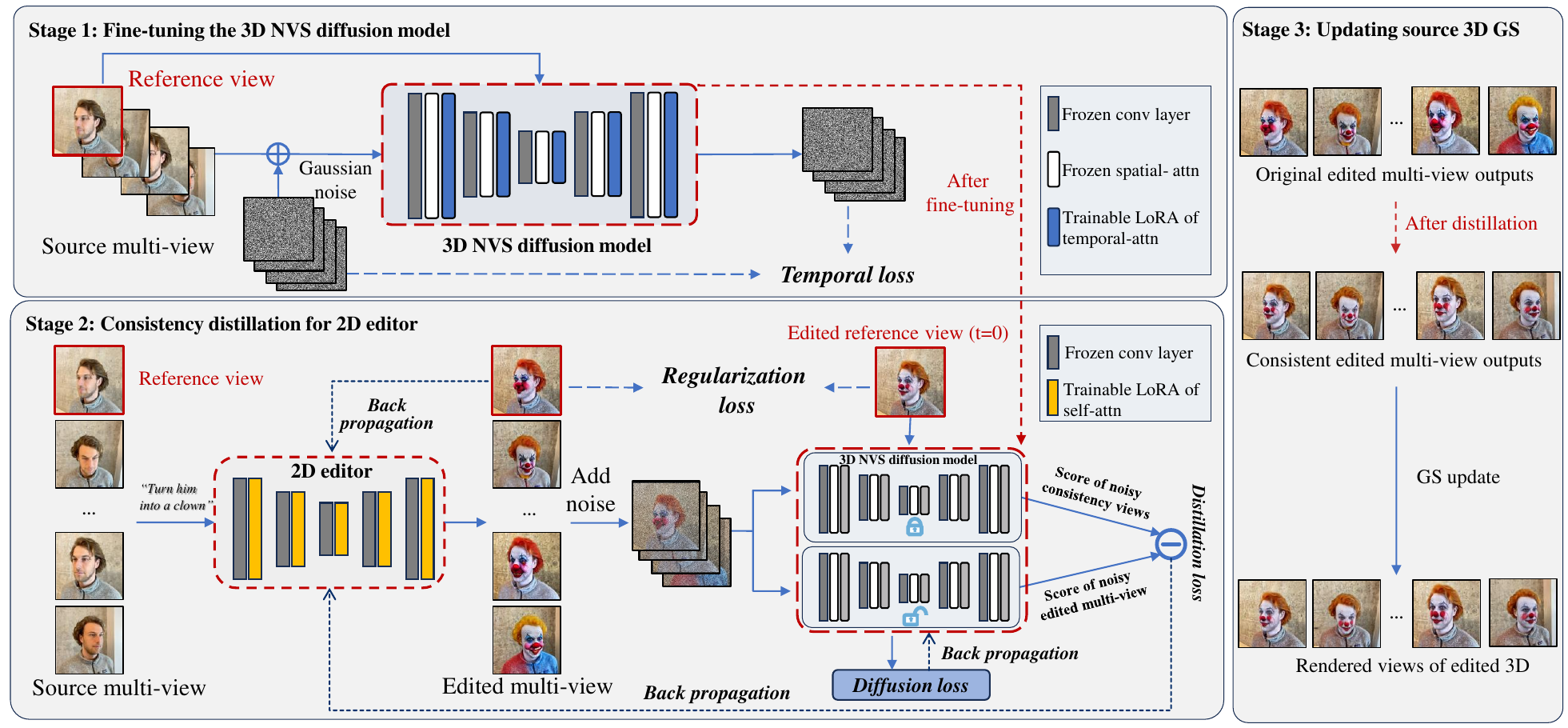}
    \caption{\textbf{Overall editing pipeline of DisCo3D.}
    (\textit{Stage 1}) \textit{Fine-tuning the 3D NVS diffusion model:}
    Given a 3D Gaussian Splatting (GS) scene, we render multi-view images under selected camera poses and fine-tune a 3D novel view synthesis (NVS) diffusion model by injecting LoRA modules into its temporal attention layers. This stage captures geometry- and texture-consistent priors from the source views along the camera trajectory.
    (\textit{Stage 2}) \textit{Consistency distillation for 2D editor:}
    To distill the 3D consistency priors into a 2D editor, we minimize the KL divergence between the consistency distribution \( p_{\text{cons}} \) and the edited multi-view distribution \( p_{\text{edit}} \). We estimate the score function of the edited distribution using a trainable replica of the pre-trained NVS model, while the pre-trained model itself parametrizes the consistency score function. Both the editor and edited score network are trained alternately. The distillation is implemented via ReFL-based sampling and a regularization loss is applied to retain the editor’s original editing capability.
    (\textit{Stage 3}) \textit{Updating Source 3D GS:}
    The consistent edited multi-view outputs from the 2D editor are used to update the 3D GS representation via $\ell_1$ and LPIPS~\cite{zhang2018unreasonable} reconstruction losses.
    }
    \label{fig:framework}
\end{figure}

DisCo3D enforces multi-view consistency via distilling 3D NVS diffusion priors into a 2D instruction-guided editor, constraining outputs for coherent edits. We render source images from a 3D GS~\cite{kerbl20233d} model under selected camera poses, followed by three core stages. First, we fine-tune a 3D NVS diffusion model on the source images to capture view consistency along camera trajectories (see Subsection~\ref{M1}). Next, we train the 2D editor using distillation and regularization losses (see Subsection~\ref{M2}). Finally, the source 3D GS model is updated using the edited multi-view outputs (see Subsection~\ref{M3}). Figure~\ref{fig:framework} illustrates our overall pipeline.

\subsection{Fine-tuning the 3D NVS diffusion model for cross-view consistency}\label{M1}
While recent diffusion-based NVS approaches~\cite{yu2024viewcrafter,sargent2024zeronvs,gao2025cat3d,liu2023zero,shriram2024realmdreamer,chung2023luciddreamer,you2024nvs} have demonstrated promising results, they often suffer from noticeable degradation in geometric consistency and texture fidelity when handling sparsely sampled camera trajectories. To address this issue and preserve both geometric and texture consistency under camera motion, we fine-tune a 3D U-Net-based NVS model by integrating LoRA modules into the temporal attention layers, following the I2VEdit~\cite{ouyang2024i2vedit} paradigm. Specifically, given multi-view input images of the source scene  \( \mathbf{I_s}= \{I_s^0, I_s^1, \dots, I_s^{N-1}\} \), we select a reference view \( I_s^{\text{ref}} \) as guidance. The source images are encoded into latent representations $\mathbf{z} = \{z_0, z_1, \dots, z_{N-1}\}$ which are then perturbed with noise to produce $\mathbf{z}_t$, where $\mathbf{z}_t = \alpha_t \mathbf{z_0} + \sigma_t \epsilon$, and $\alpha_t$ and $\sigma_t$ denote the noise schedule coefficients. We further inject \( I_s^{\text{ref}} \) through cross-attention conditioning.
Depending on the specific NVS architecture, additional conditional signals such as camera parameters may be required. We unify all such conditioning inputs into a single variable $\mathbf{c}$ for clarity.
The temporal LoRA training objective is formulated as:
\begin{equation}\label{eq:temporal_loss}
\mathcal{L}_{\text{temporal}} = \mathbb{E}_{t \sim U(0, 1), \epsilon \sim \mathcal{N}(0, \mathbf{I})} 
\left[ \| \epsilon_\theta(\mathbf{z}_t, t, \mathbf{c}, I_s^{\text{ref}}) - \epsilon \|_2^2 \right].
\end{equation}

\subsection{Consistency distillation for 2D editor}\label{M2}

Although the NVS model is fine-tuned on the source scene, generating novel views from an edited reference image remains a challenging task compared to directly performing edits in the source views. The process often results in semantic inconsistencies or texture distortions. Therefore, we utilize the fine-tuned NVS model solely as a teacher to provide 3D consistency priors.

To retain the strong single-view editing capabilities of the 2D editor while enabling cross-view consistency during multi-view adaptation, we establish a distillation framework that transfers 3D consistency priors from the fine-tuned NVS model (described in Subsection~\ref{M1}) to the diffusion-based 2D editor. During training, we initialize the student model with pre-trained weights and selectively insert LoRA modules into the self-attention layers only. This targeted modification preserves the integrity of editing instruction representations in the original cross-attention layers. To perform consistency distillation from the NVS teacher, we adopt the ReFL sampling strategy in the 2D editor (see Subsection~\ref{M2.1}). We then compute and minimize the sum of two loss terms: a distillation loss (see Subsection~\ref{M2.2}) and a regularization loss (see Subsection~\ref{M2.3}).

\begin{algorithm}
\floatname{algorithm}{Algorithm}
\renewcommand{\algorithmicrequire}{\textbf{Input:}}
\renewcommand{\algorithmicensure}{\textbf{Output:}}
\newcommand{\Comment}[1]{\hfill $\triangleright$ #1}
\footnotesize
\caption{Consistency distillation for the editing model.}
\label{alg1}
\begin{algorithmic}[1]
    \REQUIRE Source views \( \mathbf{I_s} = \{I_s^0, I_s^1, \dots, I_s^{N-1}\} \), reference view \( I_s^{\text{ref}} \), pretrained editor $E_{\theta^{(0)}}$ with LoRA in self-attention, distillation loss $\psi$, conditional input $c$, regularization loss $\mathcal{R}$ and weight $\alpha$.
    
    \STATE Initialize time steps $T$, training range $[T_1, T_2]$
    \STATE Sample $z \sim \mathcal{N}(0, I)$, compute $I_e^{\text{ref}} \leftarrow E_{\theta^{(0)}}(z, I_s^{\text{ref}})$

    \FOR{each iteration $i$}
        \STATE Sample time $t \sim \text{rand}(T_1, T_2)$
        \STATE Sample noise latent $\mathbf{z_T} \sim \mathcal{N}(0, \mathbf{I})$
        \FOR{$j = T, \dots, t+1$}
            \STATE \texttt{no grad:} $\mathbf{z_{j-1}} \leftarrow E_{\theta^{(i)}}(\mathbf{z_j}, \mathbf{I_s})$
        \ENDFOR
        \STATE \texttt{with grad:} $\mathbf{z}_{t-1} \leftarrow E_{\theta^{(i)}}(\mathbf{z}_t, \mathbf{I_s})$ 
        \STATE $\mathbf{z_0} \leftarrow \mathbf{z_{t-1}}$ \hfill \Comment{Predict the original latent by noise scheduler}
        \STATE $\mathbf{I_e}^{(i)} \leftarrow \mathbf{z_0}$ \hfill \Comment{From latent to edited view set}
        \STATE Select $I_e^{\text{ref}(i)}$ from $\mathbf{I_e}^{(i)}$ 
        \STATE $\mathcal{L}_{\text{distill}} \leftarrow \psi(\mathbf{I_e}^{(i)}, c)$ \hfill \Comment{Distillation loss}
        \STATE $\mathcal{L}_{\text{reg}} \leftarrow \alpha \cdot \mathcal{R}(I_e^{\text{ref}(i)}, I_e^{\text{ref}})$ \hfill \Comment{Regularization loss } 
        \STATE Update $\theta^{(i+1)} \leftarrow \theta^{(i)}$ using gradients
    \ENDFOR
    \RETURN Trained editor $E_{\theta}$.
\end{algorithmic}
\end{algorithm}

\subsubsection{ReFL sample}\label{M2.1}
We adopt the EDM noise schedule~\cite{karras2022elucidating} during the sampling process of the editor, which facilitates high-quality results with a reduced number of denoising steps. The denoising process edits the multi-view of source scene \(\mathbf{ I_s} = \{I_s^0, I_s^1, \dots, I_s^{N-1}\} \) into edited multi-view set \( \mathbf{I_e}= \{I_e^0, I_e^1, \dots, I_e^{N-1}\}\). The loss computation and gradient backpropagation are performed based on \( \mathbf{I_e} \). However, directly computing gradients on the multi-step denoised results is computationally expensive and impractical. To overcome this limitation, we implement the ReFL method from ImageReward~\cite{xu2023imagereward}, where we randomly select a latter step \( t \) (in our case, \( t \in [15, 20] \)) during the denoising process and compute gradients only for the final denoising step. The detailed implementation is provided in Algorithm~\ref{alg1}.

\subsubsection{Distillation loss}\label{M2.2}
Some works adopt the Variational Score Distillation (VSD)~\cite{wang2023prolificdreamer,yin2024one,yin2025improved} approach, resulting in specialized generators with capabilities comparable to the teacher model. Similarly, we follow this line of research and conduct consistency distillation from our 3D teacher model to the 2D editor. The objective is to minimize the KL divergence between the consistency distribution \( p_{\text{cons}} \) (defined by the teacher) and the edited multi-view distribution \( p_{\text{edit}} \) (produced by the student editor \( E_{\theta} \)):
\begin{equation}\label{eq:kl_div}
\mathcal{L}_{\text{distill}}(\theta) = D_{\text{KL}}(p_{\text{edit}}(\mathbf{I_e}) || p_{\text{cons}}(\mathbf{I_e})),
\end{equation}
where \( \mathbf{I_e} = E_{\theta} (\mathbf{z}, \mathbf{I_s}) \) represents the edited multi-view images generated from source inputs \(\mathbf{ I_s}\) and latent noise \(\mathbf{z}\).

To optimize this objective, we derive the gradient with respect to the editor parameters $\theta$. Following the derivation in VSD~\cite{wang2023prolificdreamer}, the gradient is formulated as:
\begin{equation}\label{eq:grad_score}
    \nabla_{\theta} \mathcal{L}_{\text{distill}} =
\underset{\substack{\mathbf{z} \sim \mathcal{N} (0, \mathbf{I}) \\ \mathbf{I_e} = E_{\theta} (\mathbf{z},\mathbf{ I_s})}} 
{\mathbb{E}} \!\left[ - \left(   s_{\text{cons}}(\mathbf{I_e}) - s_{\text{edit}}(\mathbf{I_e}) \right)\nabla_{\theta}E_{\theta} (\mathbf{z},\mathbf{I_s}) \right],
\end{equation} 
where \( s_{\text{cons}}(\mathbf{I_e}) \!=\! \nabla_{\mathbf{I_e}} \log p_{\text{cons}}(\mathbf{I_e}) \) and \( s_{\text{edit}}(\mathbf{I_e}) \!=\! \nabla_{\mathbf{I_e}} \log p_{\text{edit}}(\mathbf{I_e}) \) represent the score functions of the teacher and student distributions, respectively.

Directly computing these score functions is intractable. However, based on the score matching theory~\cite{song2020score,song2019generative}, the score function of a data distribution perturbed by Gaussian noise can be approximated by a diffusion model $\epsilon$. Specifically, for a noisy input $\mathbf{I_e}^{(t)} = \alpha_t \mathbf{I_e} + \sigma_t \epsilon$ at timestep $t$, the relation is:
\begin{equation}\label{eq:score_approx}
    s(\mathbf{I_e}^{(t)}, t) \approx -\frac{\epsilon(\mathbf{I_e}^{(t)}, t)}{\sigma_t}.
\end{equation}

We employ two diffusion models to estimate these scores:
\begin{enumerate}
    \item Teacher score ($s_{\text{cons}}$): We use the fine-tuned NVS diffusion model (frozen), denoted as \(\epsilon_{\text{fine-tuned}}\), to represent the geometric and texture consistency priors of the 3D scene.
    \item Student score ($s_{\text{edit}}$): We train a separate parameterized denoiser $\phi$, denoted as \(\epsilon_\phi\), to estimate the score of the current editor's distribution.
\end{enumerate}

Before training begins, we edit the reference view \( I_s^{\text{ref}} \) using the initialized editor to obtain the output \( I_e^{\text{ref}} \). Based on the specific architecture of the NVS model, we combine \( I_e^{\text{ref}} \) with the corresponding camera parameters to generate the conditional signal, denoted as \( \mathbf{c_e} \). 

To train the student score estimator $\epsilon_\phi$, we minimize the standard diffusion loss on the noisy outputs generated by the current editor:
\begin{equation}\label{eq:phi_loss}
\mathcal{L}_{\text{score}}(\phi) = \mathbb{E}_{t, \epsilon, \mathbf{z}} \left[ \| \epsilon_\phi(\mathbf{z_e}^{(t)}, t, \mathbf{c_e}, I_e^{\text{ref}}) - \epsilon \|_2^2 \right].
\end{equation}

Finally, by substituting Eq.~\eqref{eq:score_approx} into Eq.~\eqref{eq:grad_score}, we obtain the update rule for the editor parameters $\theta$. The intractable score difference is replaced by the difference between the teacher's predicted noise and the student's predicted noise:
\begin{equation}
\nabla_\theta \mathcal{L}_{\text{distill}}(\theta) \triangleq \mathbb{E}_{t, \epsilon} \big[ \omega(t) \left( \epsilon_{\text{fine-tuned}}(\mathbf{z_e}^{(t)}, t, \mathbf{c_e}, I_e^{\text{ref}}) \right. \left. - \epsilon_\phi(\mathbf{z_e}^{(t)}, t, \mathbf{c_e}, I_e^{\text{ref}}) \right) \nabla_\theta E_\theta(\mathbf{z}, \mathbf{I_s}) \big],
\end{equation}
where $\omega(t)$ is a weighting function dependent on $\sigma_t$ and the noise schedule. We sample $t \sim U(0.02T, 0.98T)$, consistent with DreamFusion~\cite{poole2022dreamfusion}. While Classifier-Free Guidance (CFG)~\cite{ho2021classifier} is utilized to establish the consistency score metric, it is explicitly omitted during score computation for multi-view edited outputs.

\subsubsection{Regularization loss}\label{M2.3}
The distillation loss encourages the original editor to produce more consistent edited multi-view outputs. However, since the teacher model is fitted on the source scene, the optimization direction for the editor may not always yield outputs that are both consistent and semantically faithful to the intended edit, even when conditioned on $I_e^{\text{ref}}$. Moreover, the inherent trade-off between maintaining edit semantics and enforcing scene consistency can result in blurry outputs and weaken the editor’s original editing capability.

To address this issue, we introduce a simple yet effective regularization loss during the update of the editor parameters $\theta$, ensuring that the distillation process does not degrade the editor’s editing ability or produce undesired blurriness. Specifically, we first store the output $I_e^{\text{ref}}$ generated by applying the editor (with initial parameters $\theta$) to a reference view $I_s^{\text{ref}}$. As training proceeds and the editor is updated from $\theta$ to $\theta'$, the same reference input produces a new output $I_e^{\text{ref}'}$. To preserve the editing behavior over time, we define a regularization loss that penalizes deviation from the original output:
\begin{equation}
\mathcal{L}_{\text{reg}} = \frac{1}{N} \left\| I_e^{\text{ref}'} - I_e^{\text{ref}} \right\|_2^2,
\end{equation}
where $N = H \times W \times C$ denotes the total number of pixels across all channels of the image.

The final objective for training the editor parameters $\theta$ is given by:
\begin{equation} \label{eq:overall_loss}
    \mathcal{L}_\mathrm{train} = \mathcal{L}_\mathrm{distill} + \alpha \mathcal{L}_\mathrm{reg},
\end{equation}
where $\alpha$ is a hyper-parameter to balance the losses. For a discussion on the selection of $\alpha$, refer to Subsection~\ref{ablation}.

\subsection{Updating source 3D GS and optional masking}\label{M3}
After training the editor, we can generate multi-view images that maintain strong consistency while faithfully preserving the intended editing semantics. These views are then used to update the 3D Gaussian Splatting (GS) representation \( \mathcal{G} \) of the source scene. Specifically, during the update process, we compute both the \(\ell_1\) loss and the LPIPS~\cite{zhang2018unreasonable} loss between the rendered images and their corresponding edited views. A robust reconstruction further alleviates minor inconsistencies, ultimately yielding the edited scene representation \( \mathcal{G}_e \).  
Moreover, the trained editor synchronously produces multi-view consistency outputs without iterative refinement required by prior works~\cite{chen2024dge} during GS optimization, thereby significantly streamlining the reconstruction process.

Additionally, since we adopt the explicit 3D GS representation, similar to prior works~\cite{chen2024gaussianeditor,gomel2024diffusion,chen2024dge}, we leverage a semantic segmentation model~\cite{cen2025segment} to generate a GS mask. This enables the final GS update process to be applied exclusively to the masked regions, enabling precise editing under specific instructions.

\section{Experiments}
\subsection{Implementation details}

Our 3D NVS diffusion model is based on ViewCrafter~\cite{yu2024viewcrafter}, which itself is built upon the pretrained video model DynamiCrafter and trained on large-scale 3D scene datasets. To reduce computational overhead, we adopt a lower-resolution variant with an output resolution of $320 \times 512$. After fine-tuning on the source scene, the model provides sufficiently strong consistency priors for our distillation framework.

We follow ViewCrafter's default 25-frame clip configuration and extend the viewpoint coverage to 49 views. The first view is used as the shared reference, and the remaining views are divided into two interleaved clips. In line with ViewCrafter's training procedure, we estimate the camera poses within each clip using Dust3R~\cite{wang2024dust3r}, reconstruct a point cloud from the reference view, and render it into target views using the corresponding camera poses to generate render maps. These render maps are concatenated with the real images and fed into the NVS model as input.
During Stage 1 (fine-tuning) and Stage 2 (distillation), render maps are generated from the original reference view $I_s^{\text{ref}}$ and the edited reference view $I_e^{\text{ref}}$, respectively. Accordingly, these render maps serve as the conditional signals for the NVS model in the two stages, denoted as $\mathbf{c}$ and $\mathbf{c_e}$.

For the editing model, we adopt InstructPix2Pix~\cite{brooks2023instructpix2pix} and employ the EDM noise schedule for efficient sampling. Editing is performed clip-by-clip, and the edited views, along with their corresponding conditional signals, are then fed into the NVS model for consistency distillation. This process achieves multi-view consistency within approximately 100 training iterations. All experiments run on a single NVIDIA H800 GPU. More details are provided in the supplementary material.

\subsection{Comparisons with prior work}
\noindent \textbf{Datasets.} To comprehensively evaluate our method, we conducted experiments using 6 scenes: three from the IN2N~\cite{haque2023instruct} dataset, commonly used in prior work, and three from the Tanks-and-Temples~\cite{knapitsch2017tanks} dataset, which features large viewpoint variations and more complex 360-degree environments. This selection assesses method adaptability across both conventional and challenging large-scale scenarios. Each dataset is evaluated using 10 distinct instructions, with detailed scene descriptions and prompts provided in the supplementary material.

\noindent \textbf{Baselines.} We compare our method with recent state-of-the-art approaches: IGS2GS (the 3DGS version of IN2N~\cite{haque2023instruct}), GaussianEditor~\cite{chen2024gaussianeditor}, and DGE~\cite{chen2024dge}. The first two methods employ the iterative dataset updates (IDU) approach to edit 3D representations. In contrast, DGE, similar to our method, focuses on achieving consistent multi-view editing. However, unlike DGE, which treats multi-view images as a video and achieves consistency by leveraging key-view 
attention features during the denoising process in the editor, our method explicitly distills 3D consistency constraints for precise cross-view editing.
\begin{figure}[t]
    \centering
    \includegraphics[width=\linewidth]{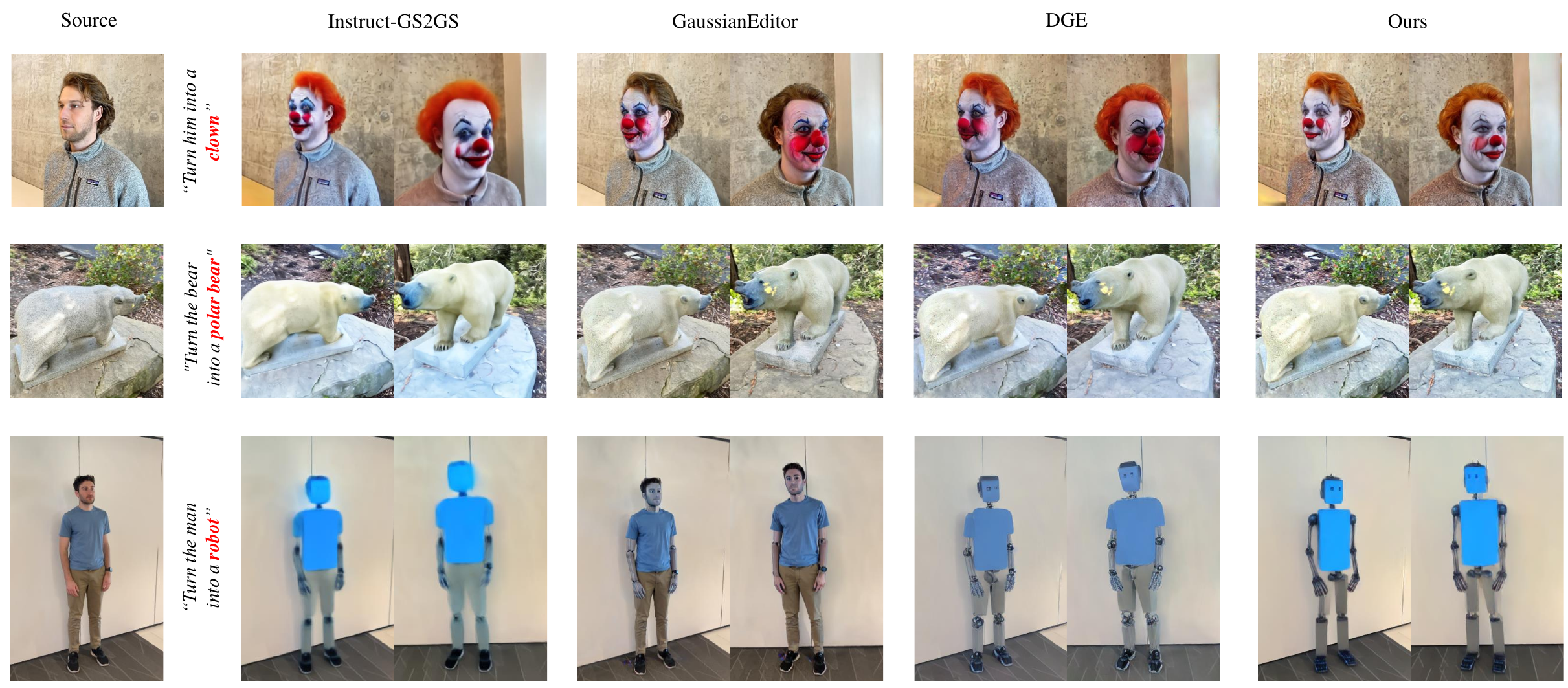}
    \caption{ \textbf{Qualitative results on IN2N dataset.} Comparison of our method with IGS2GS, GaussianEditor, and DGE in 3D editing tasks. Our method effectively preserves both fine-grained consistency and adherence to textual instructions, while other methods struggle with blurring artifacts or inconsistent multi-view edits. DisCo3D produces detailed and consistent 3D edits.}
    \label{fig:results1}
\end{figure}

\begin{figure}[t]
    \centering
    \includegraphics[width=\linewidth]{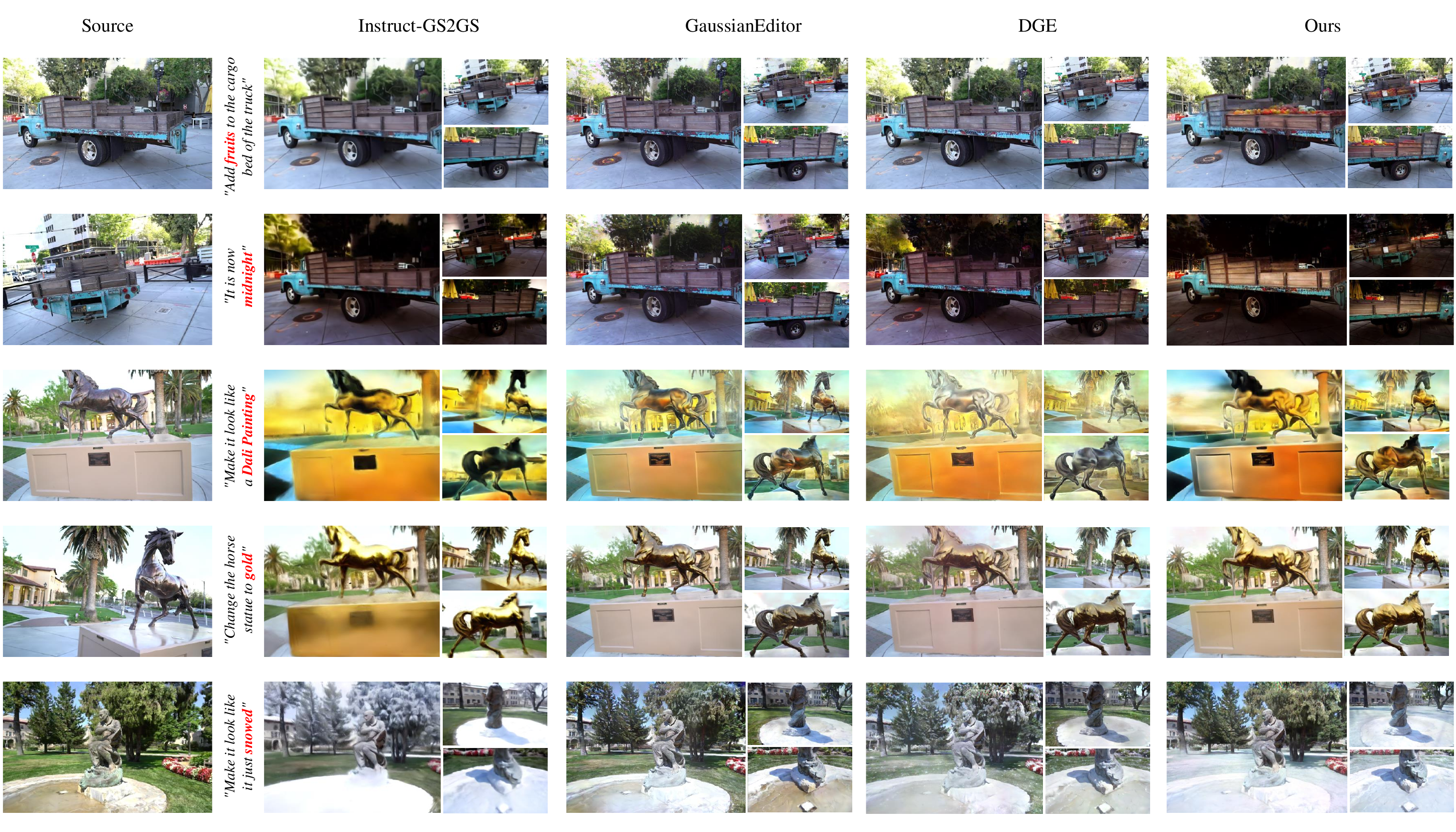}
    \caption{\textbf{Qualitative results on Tanks-and-Temples dataset.} DisCo3D demonstrates superior performance in complex 360-degree scenes, successfully executing instruction-compliant edits with precise geometric consistency (e.g., Add fruits to the cargo bed of the truck). In contrast, baseline methods exhibit critical failures, such as excessive blurring or structural inconsistencies. Our advantage stems from the 3D diffusion prior-based consistency enforcement, which holistically constrains multi-view outputs by treating all perspectives equally during distillation—unlike DGE, which relies on error-prone key-view feature attention propagation.}
    \label{fig:results2}
\end{figure}

\begin{table}[h]
    \centering
    \caption{Quantitative evaluation on IN2N and Tanks-and-Temples datasets. Higher is better. User study reflects human preference.}        
    \label{tab:quantitative_results}
    \setlength{\tabcolsep}{10pt}  
    \small
    \begin{tabular}{lcccc} 
    \toprule
    Method & 
    CLIP sim.$\uparrow$ & 
    CLIP dir. sim.$\uparrow$ & 
    CLIP dir. cons.$\uparrow$ & 
    User study$\uparrow$\\
    \midrule
    \multicolumn{5}{c}{\textbf{IN2N dataset}~\cite{haque2023instruct}} \\
    Instruct-GS2GS~\cite{haque2023instruct}     & 0.242 & 0.185 & 0.901 & 19\% \\
    GaussianEditor~\cite{chen2024gaussianeditor} & 0.231 & 0.147 & 0.857 & 25\% \\
    DGE~\cite{chen2024dge}              & \textbf{0.246} & \textbf{0.186} & 0.902 & 23\% \\
    DisCo3D (ours)                      & 0.244 & \textbf{0.186} & \textbf{0.903} & \textbf{33\%} \\
    \midrule
    \multicolumn{5}{c}{\textbf{Tanks-and-Temples dataset}~\cite{knapitsch2017tanks}} \\
    Instruct-GS2GS~\cite{haque2023instruct}     & 0.274 & 0.114 & 0.804 & 17\% \\
    GaussianEditor~\cite{chen2024gaussianeditor} & 0.269 & 0.081 & 0.728 & 13\% \\
    DGE~\cite{chen2024dge}              & 0.276 & 0.099 & 0.782 & 18\% \\
    DisCo3D (ours)                      & \textbf{0.277} & \textbf{0.118} & \textbf{0.807} & \textbf{52\%} \\
    \bottomrule
    \end{tabular}
    \vspace{0.2em}
    \begin{minipage}{\linewidth}
    \footnotesize
    \textbf{Abbreviations:} sim. = similarity; dir. sim. = directional similarity; dir. cons. = directional consistency. Bold indicates best in each metric.
    \end{minipage}
\end{table}

\noindent \textbf{Qualitative results.} 
Figure~\ref{fig:results1} demonstrates IN2N dataset comparisons. By distilling 3D consistency priors to constrain the 2D editing space, our approach achieves instruction-faithful, cross-view consistent results. In contrast, IGS2GS and GaussianEditor, which rely on the IDU strategy, often produce blurring artifacts or struggle to adhere strictly to editing instructions due to their averaging of inconsistent multi-view edits. Although DGE also attempts to enforce consistency in the editing space for more precise 3D editing, its reliance on feature-level control proves insufficient to guarantee fine-grained consistency. This limitation propagates through the reconstruction pipeline, ultimately leading to inconsistent 3D outputs. For instance, in the ``face'' editing scene, DGE generates a clown face with noticeable inconsistencies in mouth shape across views. Our method effectively addresses these limitations, achieving detailed and consistent 3D edits.

The experimental results on the Tanks-and-Temples dataset, featuring 360-degree complex scenes, are shown in Figure~\ref{fig:results2}. DisCo3D maintains robust performance even in these challenging scenarios, whereas existing methods suffer from severe limitations. IGS2GS exhibits increasing blurring artifacts as scene complexity rises, GaussianEditor frequently deviates from editing instructions due to its unconstrained 2D manipulations, and DGE, despite similar technical motivations, generates significant artifacts in geometrically intricate regions. Notably, in the task of ``Add fruits to the cargo bed of the truck'', DisCo3D is the only method that successfully completes the edit, while all baselines fail.

We attribute this superiority to fundamental differences in consistency enforcement. While DGE is extended from video editing methods by attempting to propagate denoising attention features from sparsely sampled key views, substantial content variations across 360-degree scenes weaken its multi-view constraints, and poorly edited key views may dominantly propagate artifacts. In large-scale scenes, one potential solution is to increase the number of key views. However, excessive key views introduce high computational overhead due to their mutual interactions, making the method impractical. If key views only interact with adjacent ones to reduce cost, it can easily result in incoherent or incompatible final outputs. This highlights a fundamental limitation of such propagation-based approaches. In contrast, DisCo3D ensures editing feasibility through a single reference view while enforcing equal treatment of all views during consistency distillation. This strong 3D consistency priors holistically constrain the multi-view output space, enabling stable performance even in complex scenes.

\noindent \textbf{Quantitative results.}
We adopt CLIP-based metrics~\cite{radford2021learning} for evaluation. Specifically, we compute the CLIP similarity score, CLIP directional similarity score, and CLIP directional consistency score across all training views, following the metrics reported in IN2N~\cite{haque2023instruct}.

It is important to note that editing quality is inherently subjective, and CLIP-based metrics may not fully capture fine-grained details. Therefore, we conducted a user study based on human preferences, where each question included results from our method and the baselines. Participants selected the best result from the given options, and we reported the proportion of times each method was chosen. For more details on the user study, please refer to the supplementary material.

To demonstrate adaptability across varying levels of scene complexity, we report results separately for the IN2N and Tanks-and-Temples datasets. As shown in Table~\ref{tab:quantitative_results}, on the IN2N dataset, DisCo3D achieves results comparable to the best-performing method. On the Tanks-and-Temples dataset, DisCo3D achieves the highest scores. Furthermore, in the user study, DisCo3D consistently received the highest preference scores across all methods. Notably, on the larger-scale Tanks-and-Temples dataset, participants showed an even stronger preference for DisCo3D over other methods, indicating a more substantial performance margin. This contrast is more pronounced than in the IN2N dataset, suggesting that DisCo3D is particularly effective in challenging 3D scenarios with significant viewpoint variation. Overall, these results highlight the robust adaptability and superior consistency of DisCo3D across diverse editing contexts.

Besides, for each scene editing task, the end-to-end execution of DisCo3D is almost entirely dominated by active GPU computation with negligible CPU or I/O overhead. On average, DisCo3D requires 5 minutes for fine-tuning the NVS model, followed by 7 minutes for consistency distillation, and less than 1 minute for updating the 3D Gaussian Splatting based on multi-view outputs. Notably, the fine-tuning phase is shared across all edits within the same scene and does not need to be repeated, significantly reducing computational overhead in multi-edit scenarios. As a result, our method achieves comparable execution time to IGS2GS and GaussianEditor (approximately 8 minutes), slightly higher than DGE (around 5 minutes) but delivering superior editing quality.

\subsection{Ablation study}\label{ablation}

To evaluate the impact of each component, we perform an ablation study on fine-tuning the 3D NVS diffusion model (Stage 1), consistency distillation (Stage 2), and the use of the regularization loss.

\begin{figure}[t]
    \centering
    \includegraphics[width=0.87\linewidth]{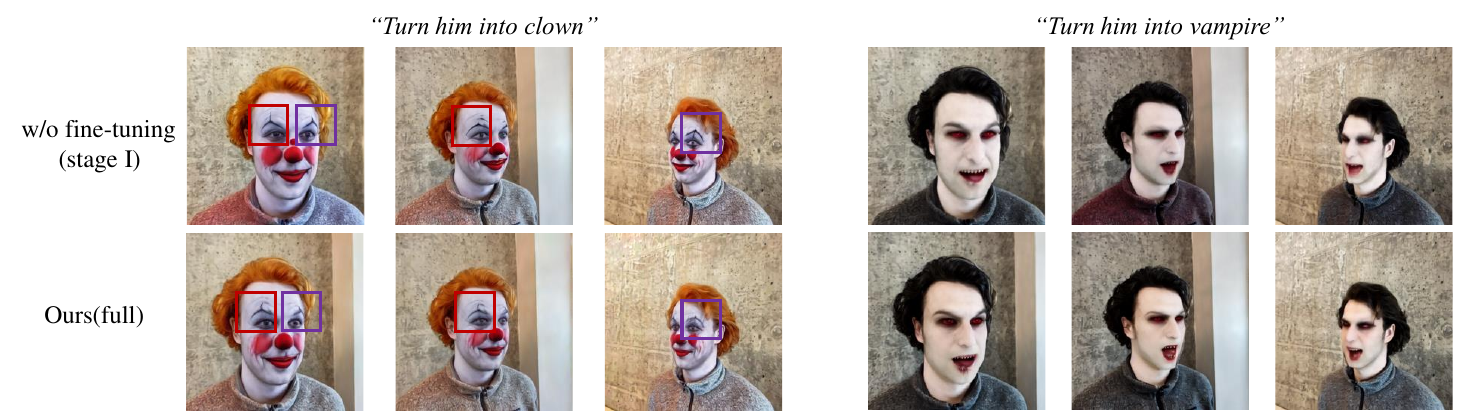}
    \caption{\textbf{Ablation on fine-tuning the NVS model:} Removing Stage 1 leads to multi-view inconsistencies (color/texture mismatch in clown eyebrows \& vampire clothing), while full implementation ensures consistent editing through stronger 3D priors.}
    \label{fig:ablation1}
\end{figure}

\begin{figure}[t]
    \centering
    \includegraphics[width=0.9\linewidth]{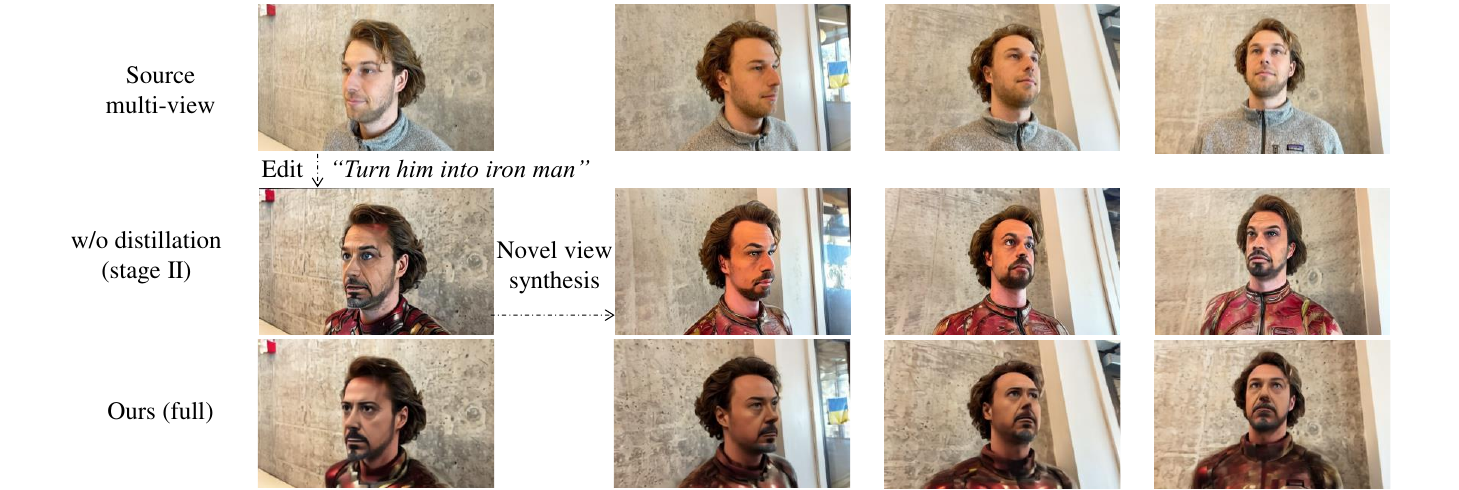}
    \caption{\textbf{Ablation on view synthesis strategy:} Replacing Stage 2 with direct novel view generation from the fine-tuned NVS model causes semantic distortions (e.g., deformed facial features), while the full pipeline preserves consistency and fidelity by distilling multi-view priors into the 2D editor.}
    \label{fig:ablation2}
\end{figure}

\noindent\textbf{Analysis on fine-tuning the 3D NVS diffusion model.}
To intuitively assess the impact of NVS model fine-tuning (Stage 1), we directly compare the outputs of the distilled editor rather than the final rendered results, keeping the number of training epochs fixed.  
Without fine-tuning the NVS model, the consistency priors provided by the 3D diffusion model are insufficient, leading to noticeable inconsistencies in texture and color across views. As shown in Figure~\ref{fig:ablation1}, omitting Stage 1 results in artifacts such as inconsistent eyebrows in the ``clown'' example (highlighted boxes), and color shifts in the clothing of the ``vampire''.  
In contrast, incorporating NVS model fine-tuning effectively resolves these issues, yielding more consistent outputs and ultimately enabling higher-quality 3D editing.

\noindent\textbf{Analysis on consistency distillation.}
The distillation stage enables the 2D editor to learn consistent multi-view edits. Since it operates directly on ground-truth views of the source scene, ensuring high fidelity and semantic alignment. In contrast, an alternative strategy that edits a reference view and generates other views via the finetuned NVS model struggles under scene-level camera variation. We tested this alternative by replacing Stage 2 with direct generation using the fine-tuned NVS model.  As shown in Figure~\ref{fig:ablation2}, this approach produces views that match the intended camera poses but results in severe facial distortions and semantic deviations—especially noticeable in the character's face. This likely stems from inherent biases in the NVS model, which remain even after fine-tuning, making it difficult to synthesize consistent novel views from edited inputs. In contrast, our full pipeline circumvents this issue by leveraging the NVS model solely to provide consistency priors, while the editor remains responsible for generating outputs that are both semantically accurate and visually coherent. As shown in Table~\ref{tab2}, incorporating Stage 2 significantly improves performance across all metrics on the ``face'' dataset.

\begin{table}[!t]
\centering
\footnotesize
\caption{Ablation on the effectiveness of the distillation stage (Stage 2).}
\label{tab2}
\tabcolsep 10pt
\begin{tabular}{lccc}
\toprule
  Settings & CLIP similarity$\uparrow$ & CLIP directional similarity$\uparrow$ &  CLIP directional consistency$\uparrow$  \\
        \midrule[1pt]
        w/o stage 2 & 0.209 & 0.140 & 0.868 \\
        Ours(full) & \textbf{0.237} & \textbf{0.190} & \textbf{0.895} 
          \\ 
\bottomrule
\end{tabular}
\end{table}

\begin{figure}[h]
    \centering
    \includegraphics[width=0.9\linewidth]{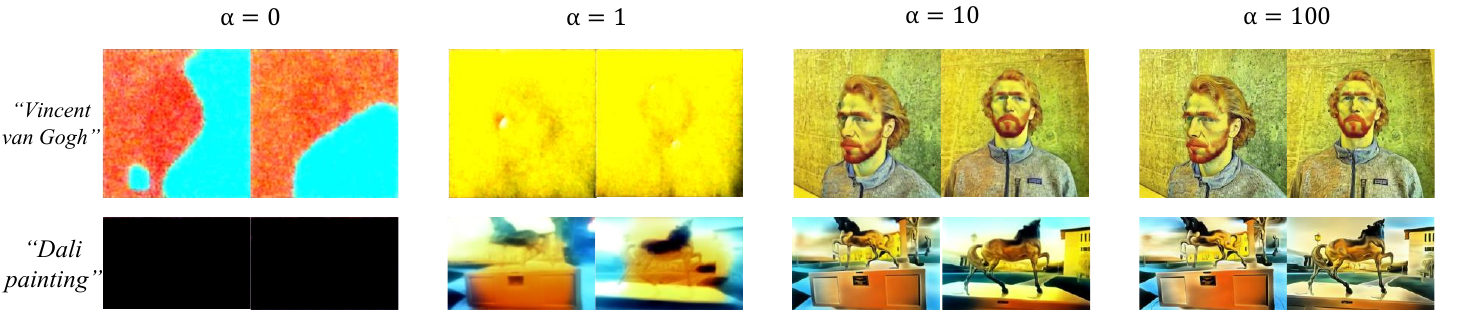}
    \caption{\textbf{Effect of regularization loss:} Setting $\alpha=0$ or $\alpha=1$ leads to semantically misaligned outputs, while overly small $\alpha$ (e.g., $\alpha=10$) results in blurred textures. Our default setting ($\alpha=100$) balances consistency optimization and fidelity preservation, enabling high-quality edits.}
    \label{fig:ablation3}
\end{figure}

\noindent\textbf{Analysis on regularization loss.} 
We further analyze the role of regularization loss and the impact of different $\alpha$ by comparing the distilled editor’s output under identical settings, varying only $\alpha$. As shown in Figure~\ref{fig:ablation3}, setting $\alpha=0$ or $\alpha=1$ leads to implausible results. This is because the absence of regularization loss during distillation degrades the capability of the 2D editor, leading to an optimized distribution that is consistent across views, but entirely misaligned with semantic information. With $\alpha=10$, the output lacks sufficient texture clarity, indicating that a weak regularization signal is still inadequate for preserving fidelity. In contrast, larger values of $\alpha$ better maintain the pretrained editor’s behavior, resulting in high-fidelity and semantically accurate outputs. Therefore, in our experiments, we set $\alpha=100$ to strike a balance between multi-view consistency and editing quality.

\begin{figure}[!ht]
    \centering
    \includegraphics[width=\linewidth]{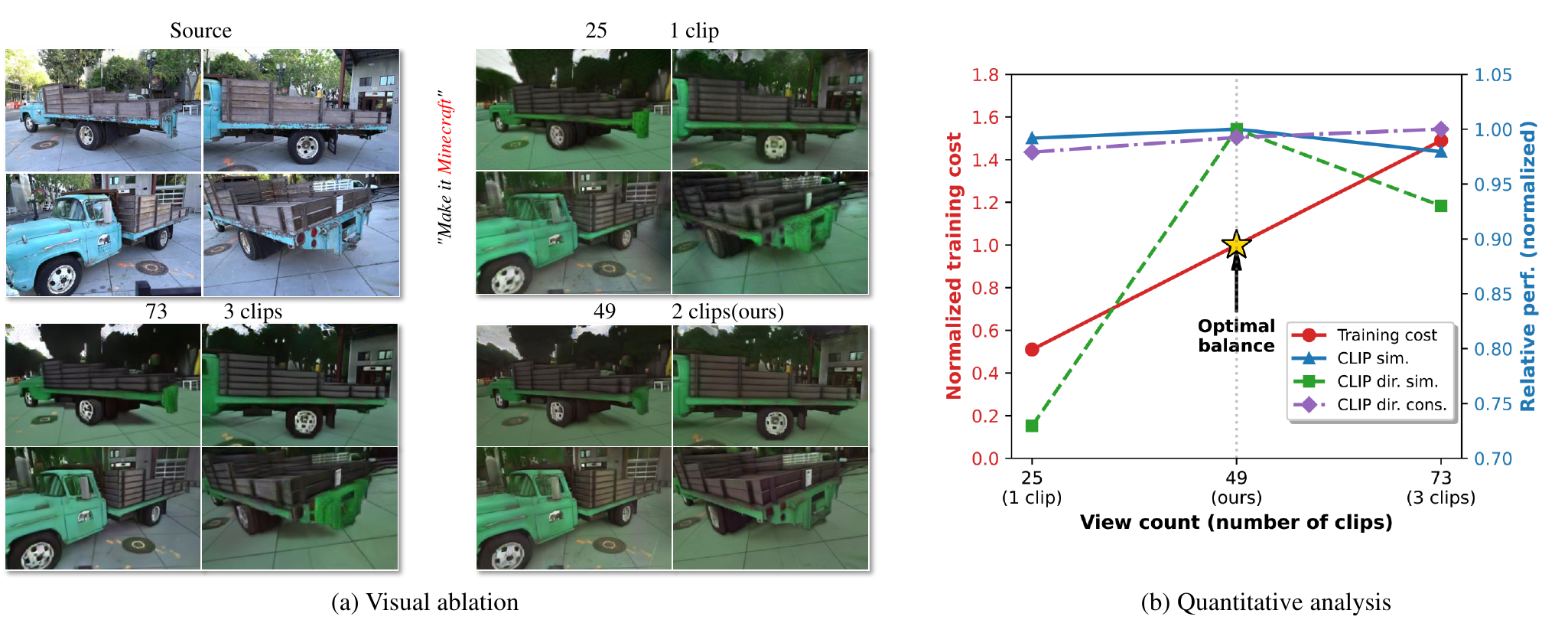}
    \caption{\textbf{Ablation study on view density:} (a) The 25-view setting leads to blurring artifacts vs. sharp details in our 49-view setting. (b) Quality metrics saturate at 49 views while cost grows linearly, confirming our \textbf{optimal balance}.}
    \label{fig:view_ablation}
\end{figure}

\noindent\textbf{Impact of view density and computational efficiency.} 
We investigate the trade-off between computational cost and generation quality by varying the number of sampled views used for score distillation. As shown in Figure~\ref{fig:view_ablation}, the computational cost scales linearly with view density. 
Qualitatively, the sparse 25-view setting ($\sim0.5\times$ cost) fails to provide sufficient consistency constraints, resulting in noticeable blurring artifacts (Figure~\ref{fig:view_ablation}(a)). 
Conversely, our default 49-view setting successfully recovers sharp, high-frequency details. 
Quantitatively, as shown in Figure~\ref{fig:view_ablation}(b), we observe that the semantic alignment metric (CLIP dir. sim.) is particularly sensitive to the view count, whereas the other evaluated metrics remain relatively stable. Interestingly, further increasing the view count to 73 ($\sim1.5\times$ cost) yields diminishing returns and even a slight decline specifically in CLIP directional similarity, likely due to the introduction of conflicting optimization noise from overly dense view sampling.
Thus, we identify the 49-view configuration as the optimal balance point between performance and efficiency.

\begin{figure}[t]
    \centering
    \begin{minipage}[c]{0.46\linewidth}
        \centering
        \includegraphics[width=\linewidth]{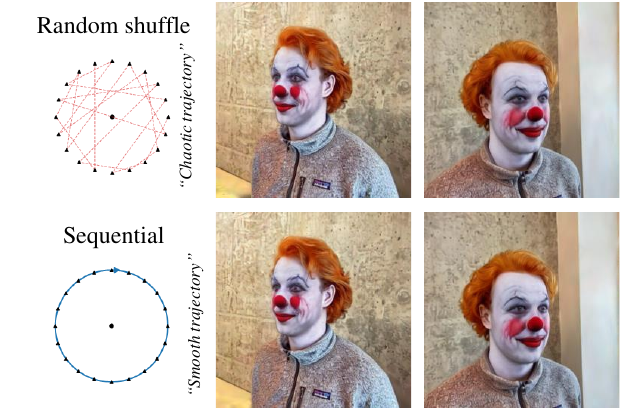}
        \vspace{-2mm}
        \captionof{figure}{\textbf{View order invariance.} Visual comparison showing that randomizing input order yields geometrically consistent results identical to the sequential input.}
        \label{fig:shuffled_ablation}
    \end{minipage}
    \hfill
    \begin{minipage}[c]{0.5\linewidth}
        \centering
        \captionof{table}{Quantitative comparison. Negligible gap confirms robustness.}
        \label{tab:order_metrics}
        \vspace{2mm} 
        \resizebox{\linewidth}{!}{
            \begin{tabular}{lccc} 
                \toprule
                Method & CLIP sim. & \begin{tabular}{@{}c@{}}CLIP dir. \\ sim.\end{tabular} & \begin{tabular}{@{}c@{}}CLIP dir. \\ cons.\end{tabular} \\
                \midrule
                Rand. shuf. & 0.257 & 0.184 & 0.916 \\
                Sequential & 0.259 & 0.187 & 0.912 \\
                \bottomrule
            \end{tabular}
        }
    \end{minipage}
\end{figure}

\noindent\textbf{Robustness to camera trajectory.} 
We further verify the sensitivity of our method to the sequential order of input views to assess its dependence on temporal smoothness. We established a baseline with a completely randomly shuffled input order, contrasting it with our standard sorted smooth trajectory. 
As presented in Figure~\ref{fig:shuffled_ablation} and Table~\ref{tab:order_metrics}, the shuffled input yields results that are quantitatively and qualitatively nearly identical to the sequential input, with a negligible performance gap. 
We attribute this permutation invariance to the strong multi-view consistency priors of the NVS teacher, which evaluates view likelihood based on pose conditioning rather than sequence history. This demonstrates that spatial coverage, rather than temporal order, is the decisive factor for our framework.

\section{Limitations} \label{sec:limitation}
While DisCo3D achieves high-fidelity consistent editing, it is subject to the limitations of its underlying foundation models. First, current NVS models often struggle in sparse-view settings. This leads to suboptimal consistency priors, which we currently compensate for through per-scene fine-tuning, introducing additional computational overhead. Second, the framework's capability is inherently bounded by the 2D diffusion editor, making it face challenges with instructions requiring significant structural changes (e.g., extensive pose deformation). If the 2D editor fails to propose valid geometric changes, our 3D distillation process cannot magically recover them. 

Furthermore, the consistency distillation process can occasionally break down under extreme occlusion. When an editing instruction dictates significant texture or semantic changes to the back of an object (which is entirely unseen in the reference view), the 2D editor tends to hallucinate inconsistent content across different novel views. If the number of views containing these hallucinated artifacts is too large, these severely conflicting signals will exceed the constraining capacity of the 3D NVS teacher. Consequently, this leads to blurry or ghosting artifacts in the previously occluded regions of the final 3D GS. Therefore, the method is currently best suited for texture editing, stylization, and moderate geometric modifications. We provide detailed visual examples and further analysis of these failure cases in the supplementary material.

\section{Conclusion}

We present DisCo3D, a novel 3D editing framework that modifies models via consistency-aware views generated by a 3D-consistent distilled editor. By constraining the editor's output space with a 3D diffusion model's consistency priors, our method ensures multi-view coherence while using a reference view to preserve semantic fidelity. This enables efficient, high-quality 3D editing without the need for cumbersome iterative refinement, making our approach highly effective and generalizable for a wide range of 3D editing tasks.

Despite its effectiveness, our method has certain boundaries that point to exciting avenues for future work. To overcome current challenges associated with extreme occlusions and significant structural deformations, future research could explore integrating explicit 3D geometric constraints or depth-aware guidance into the 2D editing pipeline to reduce hallucination. Furthermore, as foundation novel view synthesis models continue to advance, we aim to improve the robustness of our consistency priors to ultimately eliminate the reliance on per-scene fine-tuning, paving the way for a fully feed-forward, zero-shot 3D editing framework.

\Acknowledgements{This work was supported by the Fundamental and Interdisciplinary Disciplines Breakthrough Plan of the Ministry of Education of China (No. JYB2025XDXM101).}

\Supplements{Appendix.}


\end{document}